\documentclass[11pt]{article}

\usepackage[final]{acl}

\usepackage[utf8]{inputenc}
\usepackage{times}
\usepackage{latexsym}
\usepackage{enumitem}
\usepackage{csquotes}
\usepackage[T1]{fontenc}

\usepackage{microtype}

\usepackage{inconsolata}

\usepackage{graphicx}

\title{On The Conceptualization and Societal Impact of Cross-Cultural Bias}

\author{Vitthal Bhandari \\
  University of Washington / Seattle \\
  \texttt{vitthal1@uw.edu} \\}

\begin{document}
\maketitle
\begin{abstract}
Research has shown that while large language models (LLMs) can generate their responses based on cultural context, they are not perfect and tend to generalize across cultures. However, when evaluating the cultural bias of a language technology on any dataset, researchers may choose not to engage with stakeholders actually using that technology in real life, which evades the very fundamental problem they set out to address. 

Inspired by the work done by \cite{blodgett-etal-2020-language}, I set out to analyse recent literature about identifying and evaluating cultural bias in Natural Language Processing (NLP). I picked out 20 papers published in 2025 about cultural bias and came up with a set of observations to allow NLP researchers in the future to conceptualize bias concretely and evaluate its harms effectively. My aim is to advocate for a robust assessment of the societal impact of language technologies exhibiting cross-cultural bias.
\end{abstract}

\section{Introduction}

A growing body of work in NLP is shifting towards evaluating how well language models can adapt their generated responses to changes in cultural context. More recently, any unwanted generalization or brittleness in the response of a language model is deemed as cultural bias.

Given the normative nature of bias, and the far-reaching consequences of unintended harmful effects of language systems in the wild, it is important to look at these systems from the perspective of bias holistically.

In this study, I surveyed 20 papers published in 2025 about culture and bias. I probe these papers on the basis of their concreteness of the definition of bias, their identification of why bias is harmful, who is affected by that bias, in what ways it affects people, and what techniques are used to mitigate bias in language technologies.

Through this study, I contribute to the literature in the following way - 

\begin{itemize}
    \item I show that current literature in cultural bias fails to accurately define bias or the people who are directly and indirectly affected by it
    \item I show that current research lacks a comprehensive evaluation and understanding of why biased systems are actually harmful, and in what ways they manifest their harms
    \item I reveal that though researchers are using techniques outside of NLP in mitigating bias, there is a large breadth of literature to be covered, which can help us quantify and overcome this phenomenon
\end{itemize}

I argue that future research in cultural bias should concretely define what constitutes ``bias'', correctly identify who is affected by it, how it affects them, and provide reasons as to why that particular behavior is considered problematic or harmful.

\section{Related Work}

My work is primarily motivated by \citet{blodgett-etal-2020-language}, who performed a similar, much larger survey in 2020. The authors surveyed 146 papers about bias and revealed findings about definitions, motivations, and techniques used in those papers. 

\citet{gamboa-etal-2025-social} reviewed 106 papers and analyzed social bias in multilingual language models. They revealed a focus on high-resource languages and a lack of culturally specific benchmark adaptation. Noting the scarcity of mitigation research, the authors proposed an agenda prioritizing cultural transparency and linguistic inclusivity.

\citet{background} argue that the field of bias lacks a common understanding of ``culture'' which hinders progress evaluation in this increasingly important area, particularly for the safety and fairness of LLMs. To address this, they present a new taxonomy of cultural elements, which builds upon existing NLP and anthropology literature. This taxonomy aims to provide a fine-grained structure for analyzing research, including elements of social interaction often neglected in prior NLP work.

\section{Methodology}
In this section, I share the methods used to search, filter through, select, and code papers used in this study.

\subsection{Paper Collection}
I began by identifying papers across major conferences in Computer Science and Linguistics which mention the terms ``\textbf{culture}'' and/or ``\textbf{bias}'' in their titles, namely ACL, NAACL, EMNLP, COLING, ICML, ICLR, NeurIPS, AAAI, and AIES. I used 6 specific search strings to find the papers: (1) ``culture + bias'' (2) ``cultural + bias'' (3) ``culture + safety'' (4) ``cultural + safety'' (1) ``culture'' (1) ``cultural''. Search strings 1 -- 4 are used to mine the ACL Anthology\footnote{https://aclanthology.org}, whereas strings 1 -- 6 were used for searching proceedings of ICML, ICLR, NeurIPS, AAAI, and AIES. It is important to note that I constrained my search to papers published in 2025 and beyond because I wanted to survey only the most recent papers. However, the theories and arguments framing my analyses are not limited to any timeline.

\begin{table}
    \centering
    \begin{tabular}{ccc}
        \hline
        Conference & Paper Count & \\
        \hline
        ACL & 7 & \\
        NAACL & 5 & \\
        EMNLP & 2 & \\
        COLING & 1 & \\
        ICLR & 2 & \\
        NeurIPS & 2 & \\
        AIES & 1 & \\
        \hline
    \end{tabular}
    \caption{Paper counts for conferences}
    \label{tab:counts}
\end{table}

Furthermore, the aim of this study is specifically to analyze the societal impact of cross-cultural bias in NLP. Thus, the study of culture and cultural theory in NLP and the umbrella term of bias (including, but not limited to, bias on the basis of age, race, gender, or sexual orientation) is too broad and beyond the scope of this survey. I focused my study only on papers that lie at the intersection of culture and bias. 
Papers returned by the ACL Anthology 2025 search (as opposed to other conferences) were much more than those finally selected. The selection of papers was based on the relevance of the papers to the study of cultural bias in language technologies, until a sufficient count was achieved to support the analysis in our study.  Table \ref{tab:counts} lists the counts of papers selected from different conferences for this study.

Cross-cultural bias is usually studied in the context of two or more cultures. Evaluation of language technologies on Western culture only or on cultural benchmarks in English-language only is not representative of language-specific cultural information encoded across the globe. With the availability of a number of multilingual and monolingual language models for several underrepresented languages in the world, it has become easier to evaluate cross-cultural generalization of LLMs across languages. In my survey, I studied cultural bias in English and several other languages - making my analysis both multicultural and multilingual.

\begin{table}
    \centering
    \begin{tabular}{ccc}
        \hline
        Languague/culture/nationality & Paper Count & \\
        \hline
        Western culture & 9 & \\
        Arabic language/culture & 4 & \\
        African culture & 1 & \\
        Taiwanese language/culture & 1 & \\
        Global cultures & 5 & \\
        \hline
    \end{tabular}
    \caption{Paper counts by languague/culture/nationality}
    \label{tab:lng_counts}
\end{table}

Table \ref{tab:lng_counts} lists the counts of nationalities/cultures evaluated across these papers, indicating the emergent diversity in this domain.

\subsection{Paper Coding}
\label{sec:coding}
An effective analysis of the societal impacts of a language technology system (or product) should include an end-to-end perspective analysis of the possible benefits and/or harms if this system is deployed in the wild or used in a downstream application. 

Also, since bias is normative, it is important to

Based on prior work in X, Y, and Z, I coded each selected
paper based on the absence or presence of the following parameters- 

\begin{enumerate}
    \item concreteness of their definition of bias
    \item identification of direct and indirect
stakeholders
    \item possible harms of existing language
systems (if any)
    \item how these harms affect the way in which these systems are used for all stakeholders
    \item how does the proposed technique help stakeholders overcome (harmful) biases
\end{enumerate}

Collectively, these code points help explain what harm is, who the users are, how this harm negatively impacts users, and how its harmful impact affects the adoption of language technologies. 

I explain each of the code points in detail below, including sub-categories wherever possible. 

\paragraph{Concreteness of their definition of bias or unwanted behavior.} 
Categorizing a system as ``biased'' is a normative exercise because it depends on making value judgments about its behavior. Since values and norms are individualistic characteristics based on a set of beliefs, identifying bias involves varying levels of subjectivity. Different users of a system may subjectively disagree about what is considered as bias, but researchers should explicitly articulate the ethical values or social concerns that led them to label a specific system behavior as ``bias''. A concrete definition of bias helps facilitate clear downstream uses of systems. For instance, a content moderation system built by a team of engineers in the Netherlands could have a very specific modular definition of what constitutes bias in that region as opposed to a bias classifier trained by researchers in academia, which accounts for encoding diverse opinions. Neither approach is better; however, it should be clear as to what the system is flagging as bias.

\paragraph{Identification of direct and indirect
stakeholders.} 
\citet{bender2019typology} argue that language technologies built for people should place people (i.e., stakeholders) at the center and present a typology of stakeholders involved in building and deploying such systems. A person could become a stakeholder of a language technology directly or indirectly, by choice or unknowingly. I probe the papers on their identification of relevant stakeholders for the systems they built. This would enable researchers to use value sensitive design in order to analyse the task at hand.

\paragraph{Possible harms of existing language
systems (if any).} 
It is important to analyze the possible harms from bias in existing language systems, which warrants more work on understanding and mitigating cross-cultural bias. This helps us understand why a particular system is harmful to users. Quite often, it is left as an exercise for readers to hypothesize how bias in a system could harm stakeholders using that system. I argue that instead of leaving this up to the imagination of a reader, researchers should clearly demonstrate the negative impact of bias on users of a system.

\paragraph{How these harms affect the way in which these systems are used for all stakeholders.} 
Together with identifying the possible harms of existing language systems, understanding how these harms affect stakeholders' use of these systems helps draft a clear motivation for the task at hand. There might be several ways in which the inefficiency of a language system impacts its usage among a community of people. For instance, many speech-to-text models deployed in search systems of televisions and remotes for southeast-asian languages struggle with regional dialects, and hence suffer from low adoption in urban and metro areas - users prefer to type and search as opposed to searching through voice commands.

\paragraph{How does the proposed technique help stakeholders overcome (harmful) biases.} 
Not all papers propose to overcome, reduce, or mitigate bias. Generally speaking, a \textit{harm} has to be associated with the presence of a bias for it to be considered worth eliminating, which is usually not the case (and quite casually miunderstood -- not all bias is unwanted). Nonetheless, it is helpful to know whether a paper talks about reducing unwanted bias or not, and how it proposes to do that. \citet{blodgett-etal-2020-language} talk about how papers' techniques aren't well grounded in literature outside of NLP and often times, poorly matched to their motivations. Thus, it is a good exercise to analyze how papers propose to eliminate harmful behavior of existing language systems through their work.

\section{Summary of Papers}
In this section I provide short summaries of the survey papers, to later help me frame common themes across all of them.

\citet{naous-xu-2025-origin} use multilingual and monolingual language models (LMs) trained on Arabic and English and evaluate them on three tasks (text infilling, extractive QA, and NER) on a new benchmark CAMeL-2. The purpose of their evaluations is to quantify the degree of favouritism shown by LMs towards Western Culture on entities across both the English and Arabic languages. Their analysis is purely quantitative. They do not concretely define ``cultural bias'' but give examples of the kind of bias they aim to identify. They do not provide any normative reasoning for why or how this bias is harmful, nor do they explain how it impacts end users.

 \citet{dai-etal-2025-word} present CultureSteer to mitigate Western-centric bias in LLMs, evaluated via an extended Word Association Test (WAT). They define bias as stereotypical associations'' and value misalignments,'' grounding their work in psychology. While they identify human participants in data collection, they do not explicitly define downstream stakeholders or specific usage harms. Their technique is purely quantitative, using a new metric (PWR@K) to demonstrate that steering internal representations can improve cross-cultural alignment. They reveal that LLMs default to American cultural norms, but can be steered to reflect diverse cognitive patterns. 

\citet{yang-etal-2025-cultural} introduce MultiMM, a dataset for cross-cultural multimodal metaphor processing, and the SEMD model to address ``cultural skew'' in NLP. They define bias operationally as performance degradation on non-Western data, though they lack a strong normative definition. While they identify researchers as stakeholders, they do not explicitly detail harms to end-users beyond model inaccuracy. Their analysis is quantitative, utilizing sentiment embeddings grounded in cognitive linguistics. They show that Western-centric training data causes models to fail in deciphering implicit messages in Eastern cultures, proposing sentiment enrichment to bridge this gap.

 \citet{singh-etal-2025-global} introduce Global-MMLU to address the overrepresentation of Western knowledge in multilingual benchmarks. They concretely define bias through annotations of cultural, geographic, and dialectal knowledge, identifying that 28\% of MMLU is culturally sensitive. They explicitly identify global communities as stakeholders and argue that relying on translated benchmarks reinforces bias and limits model utility. Their methodology is mixed, combining quantitative evaluation with qualitative participatory research. They demonstrate that model rankings shift significantly when evaluating culturally sensitive versus agnostic data, advocating for separate evaluations to ensure equitable technology for diverse users. 

 \citet{nacar-etal-2025-towards} evaluate the Arabic MMLU benchmark, identifying critical ``cultural misalignment'' in topics like religion and morality. They explicitly define bias through metrics like Religious Sensitivity and Social Norms, linking these biases to reduced trust and acceptance among Arabic-speaking users (stakeholders). Their approach is mixed, utilizing human annotations and automated metrics to reveal that Western-centric benchmarks contain offensive content. To overcome this, they propose the ILMAAM leaderboard and inject culturally relevant topics (e.g., Islamic Ethics), grounding their work in the specific cultural and ethical contexts of the Arabic-speaking world. 

 \citet{alwajih-etal-2025-palm} introduce PALM, a dataset covering 22 Arab countries to evaluate cultural and dialectal capabilities of LLMs. They define bias as Western-centric outputs that contradict Arab values, such as suggesting alcohol consumption after prayer. They identify Arabic-speaking communities as stakeholders, arguing that cultural misalignment reduces user trust and model utility. Their analysis uses PALM to benchmark models, revealing that while closed-source models perform better, smaller open-source models struggle with cultural nuances. They propose this human-curated dataset to foster more inclusive and culturally aware language technologies.
 
 \citet{banerjee-etal-2025-navigating} present a cultural harm evaluation dataset and a preference dataset to address cultural insensitivities in LLMs. They define bias concretely as ``cultural harm,'' where models misrepresent values or identity, distinguishing this from general harm. They identify diverse cultural groups as stakeholders, warning that harms erode community identity. Their approach involves generating culturally specific adversarial prompts and using alignment methods like ORPO to mitigate harms. They demonstrate that incorporating culturally aligned feedback significantly reduces harmful outputs, arguing this is crucial for the ethical deployment of AI across diverse landscapes.

\citet{ashraf-etal-2025-arabic} introduce an Arab-region-specific safety evaluation dataset to assess LLM safeguards. They define bias through the lens of safety and cultural sensitivity, specifically noting how responses may differ based on governmental versus oppositional viewpoints. They identify Arabic-speaking users as stakeholders, noting that unsafe outputs restrict model deployment and trust. Their methodology employs a dual-perspective evaluation framework to uncover nuance in controversial topics. They find that multilingual models often fail to handle Arabic cultural contexts safely, proposing their dataset as a resource to enhance responsible AI deployment.

\citet{susanto-etal-2025-sea} present SEA-HELM, a comprehensive evaluation suite for Southeast Asian languages, addressing the scarcity of culturally representative benchmarks. They define bias as a lack of cultural and linguistic representation that can lead to ``cultural erasure'' and social harm. They explicitly identify the 700 million speakers in the SEA region as stakeholders. Their methodology is holistic, covering five pillars including linguistics, culture, and safety, and relies heavily on participatory research with native speakers. They propose this suite and a leaderboard to standardize comparisons and encourage the development of models that serve local needs.

\citet{yadav-etal-2025-beyond} explore cultural value sensitivity in Multimodal Large Language Models (MLLMs), using images as proxies for culture. They define bias as the tendency to favor specific cultural perspectives, which can misrepresent or offend diverse communities. They identify global populations as stakeholders, specifically noting disparities between high- and low-income representations. Their technique involves probing models with World Values Survey questions contextually primed with cultural images. They find that visual cues improve cultural alignment for certain topics but reveal that larger models do not necessarily guarantee better cultural sensitivity, highlighting the complexity of multimodal alignment. 

\citet{chiu-etal-2025-culturalbench} introduce CULTURALBENCH to address the uneven cultural representation in LMs, covering 45 regions including underrepresented ones like Bangladesh and Zimbabwe. They define bias as a performance gap between single-answer and multi-answer questions, identifying a ``mode-seeking'' tendency where models over-converge to stereotypes. Using a human-AI red-teaming pipeline (CulturalTeaming), they collect 1,696 verified questions. Their evaluation reveals that models struggle significantly with nuanced cultural questions (Hard set accuracy <62\% vs. Human 92\%), implying that current systems may be unhelpful for diverse stakeholders. They propose this robust benchmark to measure and drive progress toward more culturally helpful AI.

\citet{schneider-etal-2025-gimmick} present GIMMICK, a multimodal benchmark spanning 144 countries to evaluate cultural knowledge in LVLMs. They define bias as the significant performance deterioration in non-Western contexts (e.g., Sub-Saharan Africa) compared to Western ones. Grounding their work in UNESCO's Intangible Cultural Heritage data, they identify that models perform better on tangible aspects (food) than intangible ones (rituals). They argue that this lack of ``globally equitable AI'' limits model efficacy for worldwide users. Their analysis of 31 models reveals strong Western biases and shows that while model size correlates with performance, even large proprietary models struggle with nuanced cultural understanding.

\citet{feng-etal-2025-culfit} propose CulFiT, a training paradigm to mitigate the ``Western-centric perspective'' of LLMs that undermines universal equality. They identify harms such as stereotyping and language inconsistency, where models answer correctly in local languages but fail in English. They introduce a fine-grained reward modeling approach (Cultural Precision/Recall) and a multilingual critique data synthesis pipeline. Grounded in Hofstede’s cultural dimensions, their method significantly improves model performance on cultural benchmarks (e.g., GlobalCultureQA), demonstrating that targeted, multilingual fine-tuning can enhance cultural sensitivity and reduce discrimination for stakeholders in low-resource regions.

\citet{nimo-etal-2025-africa} introduce the Africa Health Check benchmark to probe cultural bias in medical LLMs, focusing on African traditional medicine. They define bias as the default to Western allopathic recommendations while omitting indigenous practices, which risks worsening health inequalities. By evaluating models on a dataset of 130+ country-remedy pairs, they quantify bias using a novel Cultural Bias Attribution (CBA) metric adapted from Integrated Gradients. They show that while prompt adaptation can reduce bias, persistent defaults to Western treatments restrict the safe deployment of these tools for African clinicians and communities, necessitating culturally informed evaluation strategies.

\citet{pham-etal-2025-cultureinstruct} introduce CULTUREINSTRUCT, a large-scale dataset of 430k synthetic instructions to reduce ``severe cultural bias'' in LLMs. They aim to ensure models operate unbiasedly across diverse contexts, covering 11 cultural domains like history and religion. Their automated pipeline filters documents from the Dolma corpus and generates instructions using a specialized model (Bonito). Evaluation on benchmarks like CANDLE and CULTURALBENCH shows that fine-tuning with their dataset consistently improves cultural knowledge and reduces linguistic bias. They argue this approach enhances the generalization potential of LLMs, making them more inclusive and fair for global stakeholders. 

\citet{Rao_Nagarajan_Venkatesan_Cherubini_Jayagopi_2025} investigate cultural bias in LLM-based hiring evaluations by analyzing transcripts from Indian and UK job seekers. They define bias as systematic scoring disparities where LLMs favor Western linguistic norms, such as lexical diversity and sentence complexity. Using anonymized transcripts and controlled identity substitutions (varying names by caste/religion), they find that while names alone do not trigger bias, linguistic styles lead to significantly lower scores for Indian candidates. This suggests that LLMs inadvertently penalize non-Western communication styles, potentially reinforcing inequality in automated hiring. They emphasize the need for community-centered evaluation frameworks to ensure fairness.

\citet{hsieh2025taiwanvqa} introduce TAIWANVQA, a benchmark for evaluating VLMs on Taiwanese cultural content. They define bias as the models' inability to reason about culturally specific visual elements, such as traditional foods and festivals, despite having general recognition capabilities. The dataset includes 2,736 images and over 5,000 questions categorized into recognition and reasoning tasks. Their evaluation reveals that while state-of-the-art models like GPT-4o perform well on recognition, they struggle with deep cultural reasoning. They propose a data augmentation strategy using synthesized dialogues to fine-tune models, demonstrating improved cultural adaptation for low-resource regions.

\citet{shen2025calm} propose CALM, a framework to endow LLMs with ``cultural self-awareness.'' They argue that existing models treat culture as static knowledge, leading to pragmatic errors and insensitivity. CALM disentangles task semantics from cultural features (explicit concepts and latent signals) and uses a contrastive window to structure them. A key innovation is the Identity Alignment Pool, which routes information through a Mixture-of-Experts based on communicative dimensions like Contextuality and Normativity. This allows the model to dynamically adapt its reasoning and self-correct cultural misalignments, significantly outperforming baselines on benchmarks like CultureAtlas and CREHate.

\citet{li2025attributing} introduce MEMOED, a framework to attribute culture-conditioned generations to pretraining data. They define bias as the tendency of models to prioritize diffuse associations'' (high-frequency, generic symbols) over culturally specific memorized associations'' due to data imbalances. By analyzing 110 cultures, they find that high-frequency cultures yield more memorized symbols, while low-frequency cultures suffer from generic, templated outputs. Their work highlights that models struggle to reliably recall diverse cultural knowledge, often defaulting to Western-centric or globally generic symbols. This attribution framework helps researchers trace the origins of bias to specific pretraining data patterns.

\citet{ananthram2025see} investigate the source of Western bias in Vision-Language Models (VLMs), defining bias as the performance disparity between Western and East Asian splits of visual tasks. They show that off-the-shelf VLMs perform significantly better on Western images. Through controlled experimentation with Llama2 and Baichuan2, they demonstrate that the language mix during text-only pre-training is a critical factor; models pre-trained with more Chinese text exhibit reduced Western bias in image understanding, even when prompted in English. This suggests that achieving multicultural vision requires multilingual foundation models, as language shapes the model's visual perspective.

\section{Findings}
The set of five broad themes used to code all the 20 papers according to the specifications in Section \ref{sec:coding} led to several common findings and sub-findings which I discuss in this section below.

\subsection{Definition of bias}

\paragraph{25\% of papers do not define bias concretely.} The first broad code point revealed that only 75\% of all papers provide a concrete definition of bias. An example of a well-defined bias source is given below - 

\begin{displayquote}
    \textit{In this study,we conduct a systematic analysis of how LLMs assess job interviews across cultural and identity dimensions. Using two datasets of interview transcripts, 100 from UK and 100 from Indian job seekers, we first examine cross-cultural differences in LLM-generated scores for hirability and related traits.} -- \citep{Rao_Nagarajan_Venkatesan_Cherubini_Jayagopi_2025}
    
\end{displayquote}

Further, only 45\% of papers give a clear normative reasoning for their choices. For instance, a clear normative reasoning for identifying bias is shown here -

\begin{displayquote}
    \textit{Persistent default to allopathic (Western) treatments in zero-shot scenarios suggest that many biases remain embedded in model training.} -- \citep{nimo-etal-2025-africa}
    
\end{displayquote}

In more recent research culture, I have observed that authors tend to provide the gist of their work in a cover-art or detailed figure right at the beginning of the paper. In the literature surrounding bias, it is common for authors to exemplify unwanted biased behavior of language systems in the cover art, instead of writing a detailed, modular definition of bias in the text. This led to my sub-question for this theme. I found that 75\% of all papers provide detailed examples of behavior that is considered unwanted and biased in a figure, as opposed to (or in addition to) describing it in text. For example -

\begin{displayquote}
    \textit{For instance, as shown in Figure 1, when prompted with ``red''..., British respondents tend to activate idiomatic expressions like \underline{see red} (anger), while Australians are more likely to mention bureaucratic metaphors such as \underline{red
    tape} (bureaucracy). This cross-cultural diver-
    gence...} -- \citep{nimo-etal-2025-africa}
    
\end{displayquote}

I was also interested in checking if authors defined bias \textit{implicitly} in the paper anywhere. I discovered that 5\% of papers have an implicit (hidden) definition of bias, and readers are expected to decode this implicit definition themselves.

\paragraph{Not all papers identify stakeholders.} As mentioned in Section \ref{sec:coding}, I looked into papers to see if they explicitly identified direct stakeholders and indirect stakeholders to identify the scope of impact of the language technologies being adopted for the task. 

I found that 60\% of papers either do not identify direct stakeholders at all, or identify them vaguely. Some papers define their primary stakeholders directly, such as -

\begin{displayquote}
    \textit{The research highlights the importance of cultural sensitivity in evaluating inclusive Arabic LLMs, fostering more widely accepted LLMs for Arabic-speaking
    communities} -- \citep{nacar-etal-2025-towards}
    
\end{displayquote}

Whereas others do not specify ``who'' exactly is going to benefit from more inclusive language technologies -

\begin{displayquote}
    \textit{This work contributes a novel methodological paradigm for enhancing cultural awareness in LLMs, advancing the
    development of more inclusive language technologies.} -- \citep{dai-etal-2025-word}
    
\end{displayquote}

Additionally, I found that only 10\% of papers distinctly identify the indirect stakeholders and 95\% of papers implicitly define indirect stakeholders in their study. 

\paragraph{Not all papers quantify the harms of existing language systems in place.} I targeted this code point from two distinct sub-categories. I probed papers for (1) absence or presence of a general analysis/mention of the harms of existing systems, and (2) absence or presence of specific examples annd/or facts which imply that existing language technologies are harmful to certain stakeholders.

I found that only 70\% of papers were intentional in their identification of the general harms from bias in existing langauge systems. In contrast, 35\% of all papers provide specific examples and instances where unwanted bias could prove harmful to stakeholders, thus implying the presence of existing harms in systems. Here is an example of a specific example which indicates that certain bias could prove harmful in some downstream tasks - 

\begin{displayquote}
    \textit{many methods assume that diverse concepts from different cultures are interchangeable (for example the Chinese erhu, a string instrument, and Western drums), potentially stripping away culturally specific nuance} -- \citep{ananthram2025see}
\end{displayquote}

\paragraph{Most papers do not explicitly evaluate the impact of systemic bias on technological adoption.} Through this line of question, I examined how harmful behavior stemming from biased systems affects the way in which these systems are used for all stakeholders -- does the paper show any significant impact on the adoption of these models/language technologies on account of their possible harms?

An example of a vague analysis of the impact of bias on how a particular technology is used by stakeholders is given below - 

\begin{displayquote}
    \textit{This limitation reduces their reliability in downstream tasks that demand genuine cultural sensitivity.} -- \citep{shen2025calm}
\end{displayquote}

I found that 80\% of papers gave vague explanations of how the harmful impact of language systems would negatively impact the way they are used by stakeholders, or limit their adoption.

\paragraph{Researchers use a variety of quantitative and qualitative techniques to help mitigate bias.}

Additionally, I found that 75\% of the techniques are purely quantitative.

One sub-question for this theme relied on probing if the papers use techniques that are well-grounded in literature outside of NLP such as sociolinguistics or sociocultural theory. I discovered that only 45\% of papers explore techniques to mitigate bias that are outside of NLP literature.

For instance, \citet{ananthram2025see} conducted a controlled study training mLLaVA variants to show that balancing the language mix during text-only pre-training (e.g., including more Chinese) significantly reduces Western bias in image understanding, even when prompting in English. They grounded their work in cognitive science research on cultural perception.

\section{Discussion}
In the previous section, I shared findings that revealed the current state of holistic societal analysis of research in cultural bias in NLP. This survey is meant to motivate future NLP researchers for evaluating bias from an ``impact to end users''-perspective.

All five of the parameters used to code the papers hold a strong correlation to the downstream impact of research on actual communities of people. Thus, it is advised that future work in the domain of cultural bias in NLP should begin with a clear definition of bias and identification of stakeholders to allow for a concrete motivation for the project. Further, evaluating and understanding the current systems' harms and their impact on technological adoption by end users should be considered as standard practice.

\section{Ethical Considerations}
The primary ethical consideration driving my work is the urgent need to improve how the NLP community identifies and mitigates the societal harms of cross-cultural bias. As language technologies are deployed globally, their potential to perpetuate stereotypes, marginalize linguistic communities, and cause tangible allocational and representational harms grows (Blodgett et al., 2020), however most research papers in the past present a vaugely defined picture of these harms. This survey functions as an ethical intervention by critically examining whether current research practices are sufficient to meet this challenge.

While this paper is a survey and not a new user-facing system, it engages directly with the spirit of proposals like Schmaltz at al. (2018), which advocate for ``AI Safety Disclosures''. The primary ethical consideration which should be taken by future researchers referring my study is to understand that a complete assessment of cultural bias must ultimately involve participatory design with the affected communities. 

Any language technology which is built for a product or for research, impacts users in some form or the other - directly or indirectly. In the case of cultural bias evaluation, researchers should be intentional in their definition and identification of bias, stakeholders, and impact. A vague acknowledgement or the negative impact of bias, for example, does not take into account its actual harm on a real-life community of people who use the particular language technology in practice, and thus evades the very fundamental problem it intended to address. 

Thus, my work provides a meta-analysis of the ethical audit of language technologies used for evaluation of cross-cultural bias in recent papers. I believe that authors building on my work should advocate for a stronger ``real-word impact statement'' in future papers on bias, much like Schmaltz at al. (2018) did for ``AI Safety Disclosures'', in order to ensure that bias-evaluation is based not only on benchmark performance metrics, but also grounded in human-centered assessment of harms and impact. Such a ``real-word impact statement'' should include information about (1) the stakeholders who will be affected by the language technology in discussion (2) the nature of effect - whether it will be positive, harmful, or neutral (3) how the technology affects the stakeholders. This holistic approach will allow for a well-rounded ethical audit of the societal ipmact of language technologies intended to study cross-cultural bias in NLP. 

\section{Conclusion}
Through this project, I conclude that current research in NLP around culture and bias fails to assess the societal impact of ubiquitous language technologies comprehensively. I call for researchers to include an intentional, thorough study of bias in their works, which clearly mentions what kind of bias is being talked about, how it affects end-users, all the possible harms, and how it leads to unwanted outcomes for the stakeholders.

\section*{Limitations}

There are some major limitations to this survey. The primary limitation is the lack of a more exhaustive study encompassing papers from past years and other conferences and workshops. A larger, more comprehensive survey would be able to highlight my point even more. Another limitation is that the scope of this project does not allow a human-centered evaluation of cultural bias using participatory designs. I feel members from other cultures should be consulted to assess the veracity of some of the claims. We will present any other limitations that may arise throughout the course of this project.

\section*{Acknowledgments}

I want to thank my peers enrolled in LING 575 for their valuable in-class collaboration and paper feedback, and my instructor for their help and timely suggestions.

\bibliography{custom,anthology_00}

@inproceedings{bender2019typology,
  title={A typology of ethical risks in language technology with an eye towards where transparent documentation can help},
  author={Bender, Emily M},
  booktitle={Future of artificial intelligence: language, ethics, technology workshop},
  volume={1},
  year={2019}
}

@article{Rao_Nagarajan_Venkatesan_Cherubini_Jayagopi_2025, title={Invisible Filters: Cultural Bias in Hiring Evaluations Using Large Language Models}, volume={8}, url={https://ojs.aaai.org/index.php/AIES/article/view/36703}, DOI={10.1609/aies.v8i3.36703}, abstractNote={Artificial Intelligence (AI) is increasingly used in hiring, with large language models (LLMs) having the potential to influence or even make hiring decisions. However, this raises pressing concerns about bias, fairness, and trust, particularly across diverse cultural contexts. Despite their growing role, few studies have systematically examined the potential biases in AI-driven hiring evaluation across cultures. In this study, we conduct a systematic analysis of how LLMs assess job interviews across cultural and identity dimensions. Using two datasets of interview transcripts, 100 from UK and 100 from Indian job seekers, we first examine cross-cultural differences in LLM-generated scores for hirability and related traits. Indian transcripts receive consistently lower scores than UK transcripts, even when they were anonymized, with disparities linked to linguistic features such as sentence complexity and lexical diversity. We then perform controlled identity substitutions (varying names by gender, caste, and region) within the Indian dataset to test for name-based bias. These substitutions do not yield statistically significant effects, indicating that names alone, when isolated from other contextual signals, may not influence LLM evaluations. Our findings underscore the importance of evaluating both linguistic and social dimensions in LLM-driven evaluations and highlight the need for culturally sensitive design and accountability in AI-assisted hiring.}, number={3}, journal={Proceedings of the AAAI/ACM Conference on AI, Ethics, and Society}, author={Rao, Pooja S. B. and Nagarajan Venkatesan, Laxminarayen and Cherubini, Mauro and Jayagopi, Dinesh Babu}, year={2025}, month={Oct.}, pages={2164-2176} }

@inproceedings{
hsieh2025taiwanvqa,
title={Taiwan{VQA}: Benchmarking and Enhancing Cultural Understanding in Vision-Language Models},
author={Hsin-Yi Hsieh and Shang-Wei Liu and Chang Chih Meng and Chien-Hua Chen and Shuo-Yueh Lin and Hung-Ju Lin and Hen-Hsen Huang and I-Chen Wu},
booktitle={The Thirty-ninth Annual Conference on Neural Information Processing Systems Datasets and Benchmarks Track},
year={2025},
url={https://openreview.net/forum?id=atofIc3x1q}
}

@inproceedings{
shen2025calm,
title={{CALM}: Culturally Self-Aware Language Models},
author={Lingzhi Shen and Xiaohao Cai and YUNFEI LONG and Imran Razzak and Guanming Chen and Shoaib Jameel},
booktitle={The Thirty-ninth Annual Conference on Neural Information Processing Systems},
year={2025},
url={https://openreview.net/forum?id=16QYhVFvrO}
}

@inproceedings{
li2025attributing,
title={Attributing Culture-Conditioned Generations to Pretraining Corpora},
author={Huihan Li and Arnav Goel and Keyu He and Xiang Ren},
booktitle={The Thirteenth International Conference on Learning Representations},
year={2025},
url={https://openreview.net/forum?id=XrsOu4KgDE}
}

@inproceedings{
ananthram2025see,
title={See It from My Perspective: How Language Affects Cultural Bias in Image Understanding},
author={Amith Ananthram and Elias Stengel-Eskin and Mohit Bansal and Kathleen McKeown},
booktitle={The Thirteenth International Conference on Learning Representations},
year={2025},
url={https://openreview.net/forum?id=Xbl6t6zxZs}
}

@article{background,
  publtype={informal},
  author={Chen Cecilia Liu and Iryna Gurevych and Anna Korhonen},
  title={Culturally Aware and Adapted NLP: A Taxonomy and a Survey of the State of the Art},
  year={2024},
  cdate={1704067200000},
  journal={CoRR},
  volume={abs/2406.03930},
  url={https://doi.org/10.48550/arXiv.2406.03930}
}

@inproceedings{dai-etal-2025-word,
    title = "From Word to World: Evaluate and Mitigate Culture Bias in {LLM}s via Word Association Test",
    author = "Dai, Xunlian  and
      Zhou, Li  and
      Wang, Benyou  and
      Li, Haizhou",
    editor = "Christodoulopoulos, Christos  and
      Chakraborty, Tanmoy  and
      Rose, Carolyn  and
      Peng, Violet",
    booktitle = "Proceedings of the 2025 Conference on Empirical Methods in Natural Language Processing",
    month = nov,
    year = "2025",
    address = "Suzhou, China",
    publisher = "Association for Computational Linguistics",
    url = "https://aclanthology.org/2025.emnlp-main.1246/",
    doi = "10.18653/v1/2025.emnlp-main.1246",
    pages = "24521--24537",
    ISBN = "979-8-89176-332-6"
}

@inproceedings{nimo-etal-2025-africa,
    title = "{A}frica Health Check: Probing Cultural Bias in Medical {LLM}s",
    author = "Nimo, Charles  and
      Liu, Shuheng  and
      Essa, Irfan  and
      Best, Michael L.",
    editor = "Christodoulopoulos, Christos  and
      Chakraborty, Tanmoy  and
      Rose, Carolyn  and
      Peng, Violet",
    booktitle = "Proceedings of the 2025 Conference on Empirical Methods in Natural Language Processing",
    month = nov,
    year = "2025",
    address = "Suzhou, China",
    publisher = "Association for Computational Linguistics",
    url = "https://aclanthology.org/2025.emnlp-main.1639/",
    doi = "10.18653/v1/2025.emnlp-main.1639",
    pages = "32207--32220",
    ISBN = "979-8-89176-332-6"
}

@inproceedings{gamboa-etal-2025-social,
    title = "Social Bias in Multilingual Language Models: A Survey",
    author = "Gamboa, Lance Calvin Lim  and
      Feng, Yue  and
      Lee, Mark G.",
    editor = "Christodoulopoulos, Christos  and
      Chakraborty, Tanmoy  and
      Rose, Carolyn  and
      Peng, Violet",
    booktitle = "Proceedings of the 2025 Conference on Empirical Methods in Natural Language Processing",
    month = nov,
    year = "2025",
    address = "Suzhou, China",
    publisher = "Association for Computational Linguistics",
    url = "https://aclanthology.org/2025.emnlp-main.1416/",
    doi = "10.18653/v1/2025.emnlp-main.1416",
    pages = "27845--27868",
    ISBN = "979-8-89176-332-6"
}

\appendix



\end{document}